\documentclass[conference]{IEEEtran}
\IEEEoverridecommandlockouts

\makeatletter
 \let\old@ps@headings\ps@headings
 \let\old@ps@IEEEtitlepagestyle\ps@IEEEtitlepagestyle
 \def\confheader#1{%
 \def\ps@IEEEtitlepagestyle{%
 \old@ps@IEEEtitlepagestyle%
 \def\@oddhead{\strut\hfill#1\hfill\strut}%
 \def\@evenhead{\strut\hfill#1\hfill\strut}%
 }%
 \ps@headings%
 }
 \makeatother

\confheader{%
 \small{This paper has been accepted to 2023 IEEE International Conference on Control, Automation and Robotics (ICCAR).}
 }

\usepackage[utf8]{inputenc}
\usepackage[T1]{fontenc}

\usepackage{amsmath} %
\usepackage{amssymb}  %
\usepackage{multirow}
\usepackage{makecell}
\usepackage{url}
\usepackage[figurename=Fig.]{caption}

\usepackage{rotating}
\usepackage{booktabs}%
\usepackage{cite}
\usepackage[textsize=tiny,final]{changes}
\usepackage[font=footnotesize]{caption}
\usepackage[font=footnotesize,skip=5pt]{caption}

\usepackage[compact]{titlesec}
\titlespacing{\section}{0pt}{2ex}{1ex}
\titlespacing{\subsection}{0pt}{1ex}{0ex}
\titlespacing{\subsubsection}{0pt}{0.5ex}{0ex}
\def\BibTeX{{\rm B\kern-.05em{\sc i\kern-.025em b}\kern-.08em
    T\kern-.1667em\lower.7ex\hbox{E}\kern-.125emX}}

\begin{document}

\title{
\LARGE \bf
{Dynamic Modeling and Validation of Soft Robotic Snake Locomotion}
\thanks{This work is supported in part by the National Science Foundation (NSF) Grants IIS-2008797, CMMI-2048142, CMMI-2133019, and CMMI-2132994.}
}

\author{\IEEEauthorblockN{Dimuthu D. K. Arachchige, Sanjaya Mallikarachchi, Iyad Kanj}
\IEEEauthorblockA{
\textit{School of Computing, Jarvis College of Computing \& Digital Media,} \\
\textit{DePaul University},
\textit{Chicago, IL, USA} \\
\{darachch, smallika, ikanj\}@depaul.edu}
\and
\IEEEauthorblockN{Dulanjana M. Perera}
\IEEEauthorblockA{
\textit{Dept. of Multidisciplinary Engineering,} \\
\textit{Texas A\&M University},
\textit{College Station, TX, USA} \\
dperera@tamu.edu}
\and
\IEEEauthorblockN{Yue Chen}
\IEEEauthorblockA{
\textit{Dept. of Biomedical Eng.,} \\
\textit{Georgia Institute of} \\
\textit{Technology, Atlanta, GA, USA} \\
yue.chen@bme.gatech.edu}
\and
\IEEEauthorblockN{Hunter B. Gilbert}
\IEEEauthorblockA{
\textit{Dept. of Mechanical \& Industrial Eng.,} \\
\textit{Louisiana S. University},
\textit{Louisiana, USA} \\
hbgilbert@lsu.edu
}
\and
 \IEEEauthorblockN{Isuru S. Godage}
 \IEEEauthorblockA{
 \textit{Dept. of Eng. Tech. \& Industrial Distribution} \\
  \textit{and J. Mike Walker \textquotesingle66 Dept. of Mechanical Eng.,} \\
 \textit{Texas A\&M University},
 \textit{College Station, TX, USA} \\
 igodage@tamu.edu}
}

\maketitle

\begin{abstract}
		Soft robotic snakes made of compliant materials can continuously deform their bodies and, therefore, mimic the biological snakes' flexible and agile locomotion gaits better than their rigid-bodied counterparts. Without wheel support, to date, soft robotic snakes are limited to emulating planar locomotion gaits, which are derived via kinematic modeling and tested on robotic prototypes.  Given that the snake locomotion results from the reaction forces due to the distributed contact between their skin and the ground, it is essential to investigate the locomotion gaits through efficient dynamic models capable of accommodating distributed contact forces.  We present a complete spatial dynamic model that utilizes a floating-base kinematic model with distributed contact dynamics for a pneumatically powered soft robotic snake.  We numerically evaluate the feasibility of the planar and spatial rolling gaits utilizing the proposed model and experimentally validate the corresponding locomotion gait trajectories on a soft robotic snake prototype. We qualitatively and quantitatively compare the numerical and experimental results which confirm the validity of the proposed dynamic model.  
\end{abstract}

\section{Introduction\label{sec:Introduction}}

	Snakes’ unique physical structure with spatial bending capabilities enables them to overcome numerous challenges in their habitats. 
	They frequently use lateral undulation, sidewinding, rectilinear, and concertina locomotion gaits to navigate terrains. Additionally, snakes use rolling gaits for multiple purposes such as climbing trees efficiently, moving their bodies sideways quickly on uneven terrains and slopes, crossing over obstacles, handling prey, fighting, etc. Over the years, roboticists have developed various robotic snake prototypes to harness these unique characteristics~\cite{yang2022snake}. 
	Compared to traditional rigid-bodied robotic snakes, the continuous bending capability of soft robotic snakes (SRSs) increases their adaptability and flexibility with the environment. 
	
	Snake locomotion results from differential friction reaction forces between the skin and the contact surface (friction anisotropy)~\cite{hu2009mechanics}. Therefore, to accurately emulate snake locomotion, it is essential to study their locomotion gaits through dynamic models that include anisotropic frictional forces. Existing snake robots lack dynamic models that can accurately and efficiently mimic seamless snake locomotion. 
	\begin{figure}[t] 
		\centering
		\includegraphics[width=0.99\linewidth]{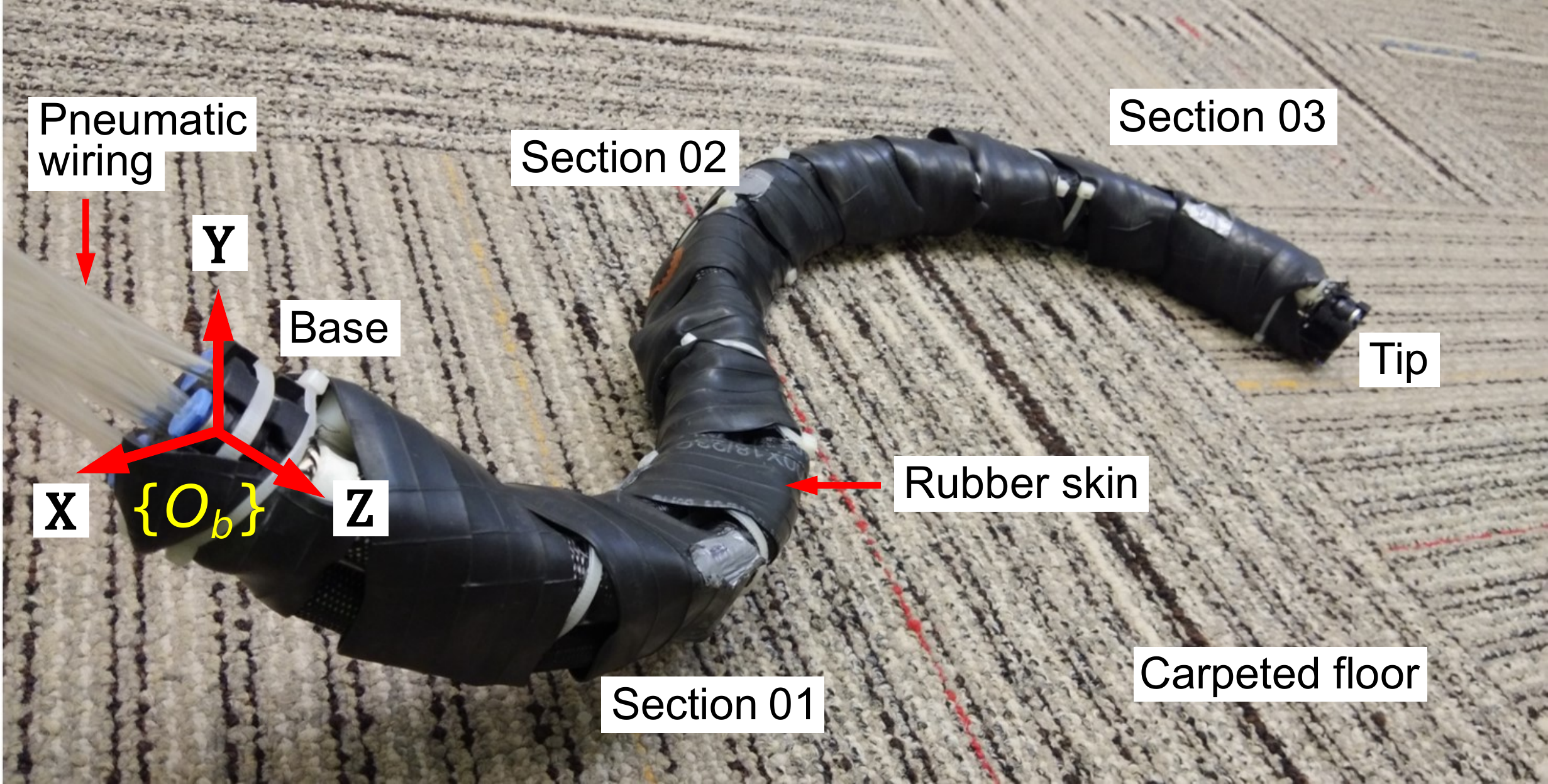}
		\caption{Soft robotic snake prototype lying on a carpeted floor.}
		\label{fig:Fig1_IntroductionImage} 
	\end{figure}	%

	Generating dynamic models for continuum robots is difficult due to the significant deformation they undergo~\cite{gilbert2019validation,godage2019center,habibi2018modelling,zheng2013model}.
	Researchers have proposed several dynamic modeling techniques for continuum robots over the years~\cite{armanini2023soft}. 
	Lumped parameter models such as those reported in~\cite{habibi2020lumped} used piecewise-constant curvature approximation of robot shape. Those models did not account for axial deformation and become invalid when the robot is subjected to complex external loading. 
	The discrete Cosserat approach~\cite{janabi2021cosserat} and finite element methods~\cite{grazioso2019geometrically} have also been used to derive the dynamics of multisection continuum robots. Their use is limited in real-time applications since they involve computationally expensive calculations.

	To date, many dynamic models for rigid robotic snakes have been proposed~\cite{girija2022survey}. The rigid robotic snakes made of discrete rigid units inherently lack continuous skin, hence their dynamic models do not reflect organic snake locomotion.
	
	Prior work on dynamic modeling of SRSs includes~\cite{luo2014theoretical,luo2015refined,luo2015slithering}.
	Their work is limited to wheeled SRSs and present planar locomotion dynamics of segmented SRSs. Compared to~\cite{luo2014theoretical}, dynamic models in~\cite{luo2015refined} and~\cite{luo2015slithering} incorporated anisotropic friction forces acting on SRS wheels and described planar locomotion. But, wheeled SRS dynamic models offer limited utility when imitating spatial SRS locomotion gaits with distributed skin-ground contact forces. However, no spatial dynamic models for SRS that support distributed contact models exist. 
	
	In our previous work~\cite{arachchige2021soft}, we proposed a pneumatic muscle actuator (PMA) powered SRS. We showed the utility of spatial bending to derive SRS locomotion gaits without wheels through planar rolling. 
	Therein, the jointspace trajectories (length variation of PMAs) for rolling locomotion derived via the complete kinematic model were directly tested on the SRS prototype on a trial-and-error basis. In this work, we extend the proposed planar rolling approach to validate spatial rolling gaits on SRSs. Hence, a spatial dynamic model of the SRS is beneficial to validate the gait performance. 
	
	Godage et al. in~\cite{godage2011shape,godage2011dynamics,godage2016dynamics} proposed dynamic modeling for variable-length continuum arms based on an integral Lagrangian approach. 
	In~\cite{godage2016dynamics}, they proposed a new spatial dynamic model for multi-section continuum arms and validated using a pneumatically actuated prototype. 
	Yet, it did not include contact modeling.
	In this work, we modify the dynamic model in~\cite{godage2016dynamics} by adding contact dynamics to accommodate SRS locomotion. Extending~\cite{godage2016dynamics}, in this work, we, i) present a complete spatial dynamic model with contact dynamics for SRSs, ii) evaluate the model in a simulation environment, and iii) validate the model on an SRS prototype for planar and spatial rolling gaits.	
	
	\section{Kinematic Modeling \label{sec:Kinematic-Modeling}}
	
	\subsection{Prototype Description\label{subsec:Prototype-Description}}
	
	\begin{figure}[tb] 
		\centering
		\includegraphics[width=1\linewidth]{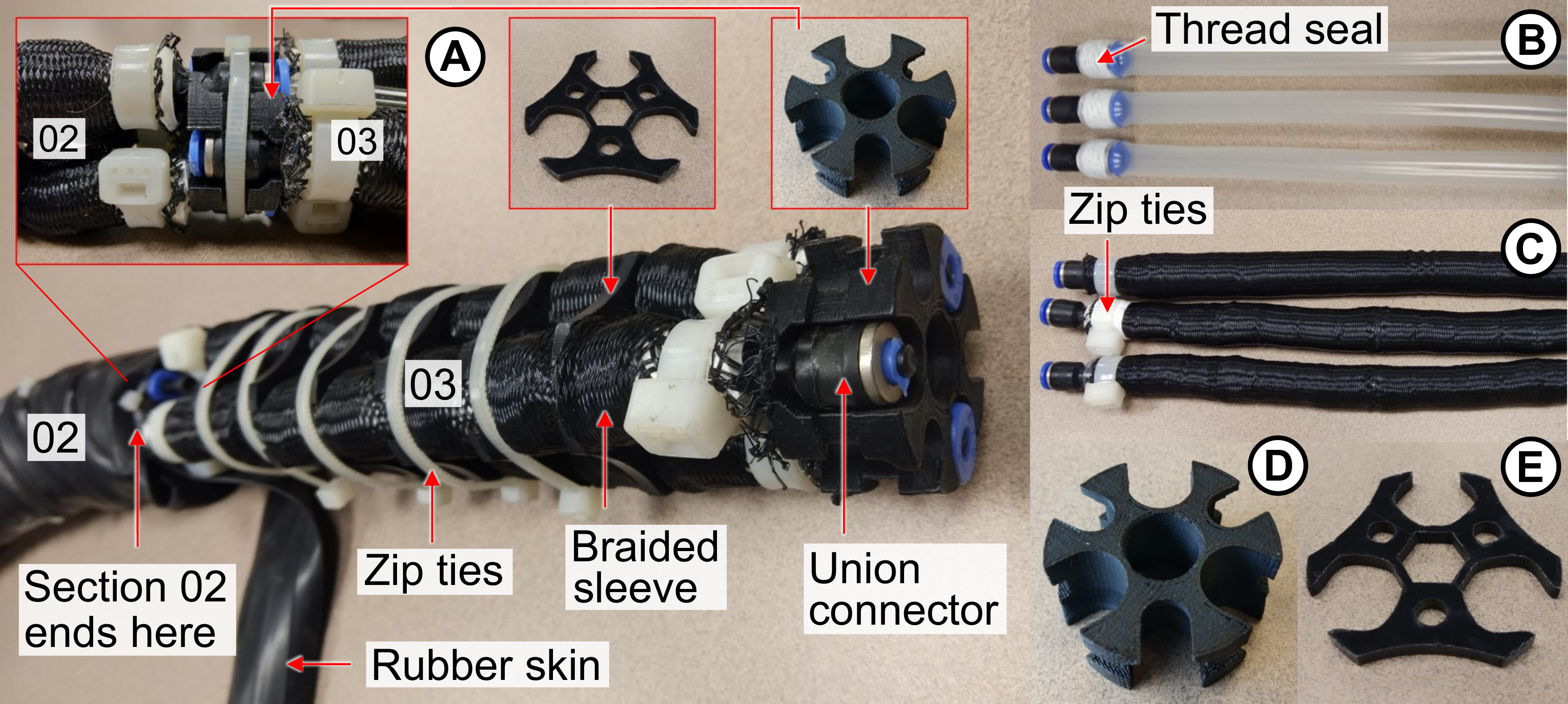}
		\caption{(A) SRS prototype -- serially arranged soft bending section assembly without rubber skin. (B) Silicone tubes of PMAs, (C) Adding Nylon braided mesh on Silicone tubes to fabricate PMAs, (D) Constrainer plates to maintain PMA spacing, (E) Mounting frames (end-plates).}
		\label{fig:Fig2_PartDesign} 
	\end{figure}
	
	The SRS prototype shown in Fig.~\ref{fig:Fig1_IntroductionImage} is assembled using three serially attached soft bending units (i.e. sections) shown in Fig.~\ref{fig:Fig2_PartDesign}A.
	An SRS section is actuated by three McKibben-type extension-mode PMAs~\cite{arachchige2021novel,amaya2021evaluation}. PMAs are fabricated using commercially available Silicone tubes, pneumatic union connectors, Nylon braided sleeves, and high strength Polyethylene fasteners (Figs.~\ref{fig:Fig2_PartDesign}B and~\ref{fig:Fig2_PartDesign}C)~\cite{arachchige2022hybrid}. Within bending sections, PMAs are mounted tri-symmetrically at a $\frac{\pi }{3}$
	angle from each other and $0.0125~m$ radius from its centerline using 3D-printed mounting frames at either end (Fig.~\ref{fig:Fig2_PartDesign}D). A PMA axially extends upon pressurizing and can sustain pressures up to $4~bars$. The unactuated length of a PMA is $0.15~m$ and can extend 50\% at $4~bars$. 
	The $0.0025~m$ thick, laser-cut Delrin constrainer plates (Fig.~\ref{fig:Fig2_PartDesign}E) help maintain PMA clearance to the central axis of a section as well as adjacent PMAs during operation without torsion and buckling.
	Further, hollowed symmetrical design of bending sections facilitates the routing of pneumatic supply lines within the robot structure.
	When pressurized, the pressure difference of PMAs generates a torque imbalance at the mounting end plates (Fig.~\ref{fig:Fig2_PartDesign}D). Based on the induced pressure differential, this torque facilitates the omnidirectional bending deformation of SRS bending units. 
	We assemble the SRS by serially connecting the bending sections via mounting frames with a $\frac{\pi}{3}$ angular offset (enlarged image in Fig.~\ref{fig:Fig2_PartDesign}A). 
	This angular offset allows the pneumatic tube routing
	without impeding the functionality of adjacent bending sections. 
	We use a uniform rubber skin (Fig.~\ref{fig:Fig2_PartDesign}A) to wrap the outer surface of the SRS to form a continuous skin-like layer for achieving uniform friction during operation. The unactuated SRS prototype is $0.60~m$ in 
	length 
	and weighs $0.35~kg$. As there are 9, independently controlled PMAs,
	the SRS has 9 actuated degrees of freedom (DoF).
	
	\begin{figure}[tb] 
		\centering
		\includegraphics[width=1\linewidth]{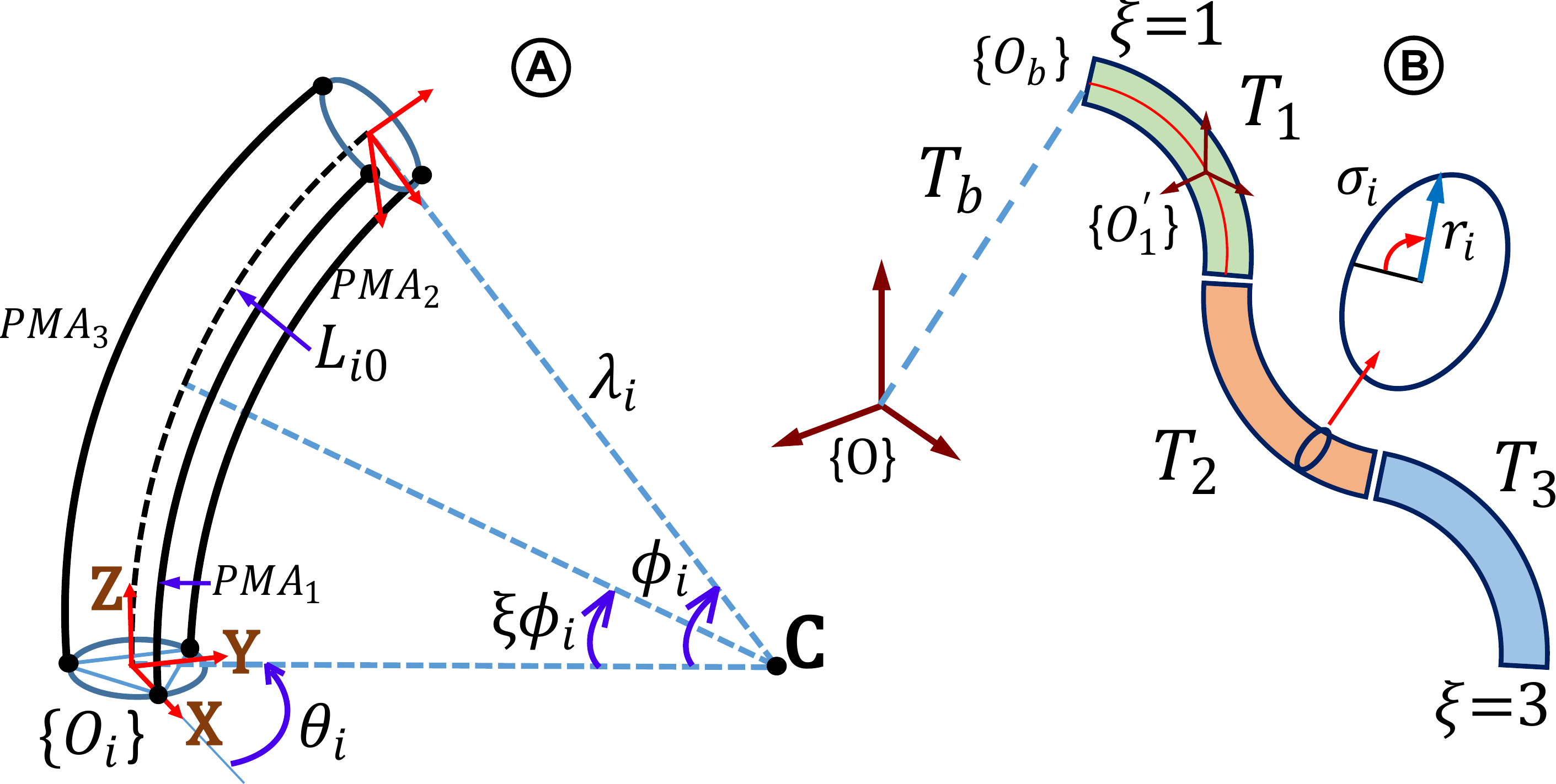}
		\caption{ SRS schematics illustrating (A) PMA arrangement of a bending section and (B) 3-section robot.}
		\label{fig:Fig3_SchematicDiagram} 
	\end{figure}
	
	\begin{figure}[tb] 
		\centering
		\includegraphics[width=1\linewidth]{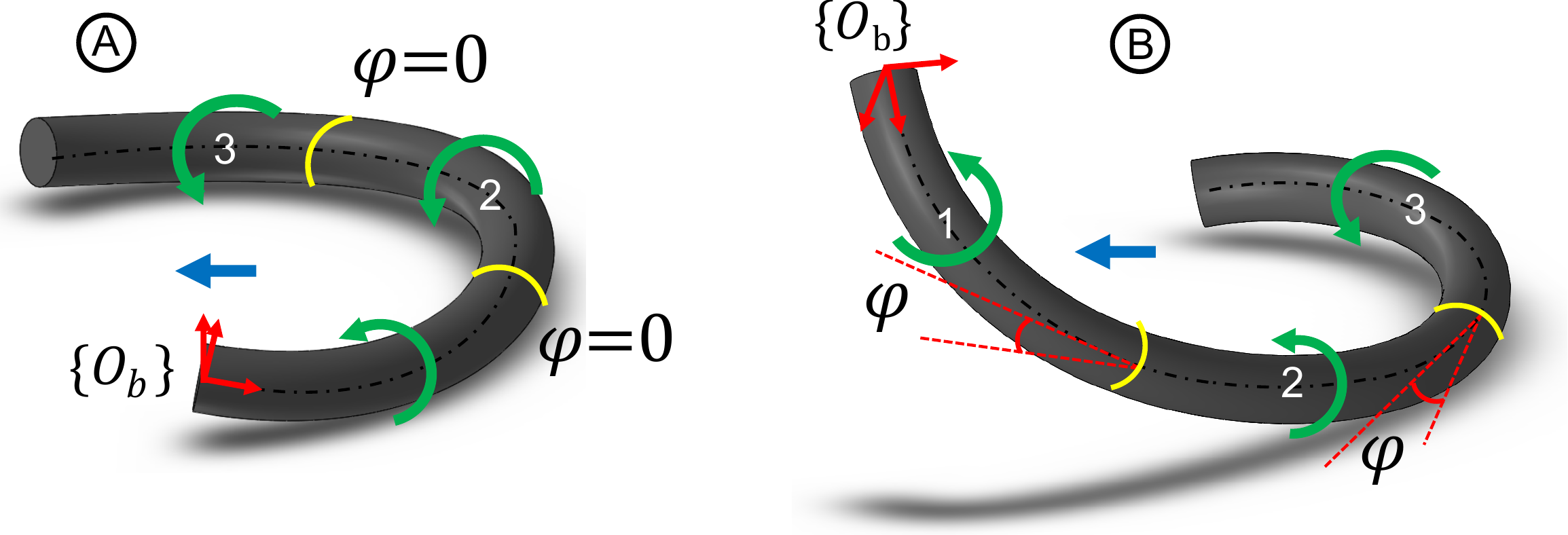}
		\caption{Gait visualization -- (A) Planar and (B) Spatial rolling.}
		\label{fig:Fig4_TrajectoryGeneration} 
	\end{figure}

	\subsection{Robot Kinematic Model\label{subsec:SRS-Kinematic-Model}} 
	
	In our previous work~\cite{arachchige2021soft}, we derived the model kinematics of the 3-section SRS along its neutral axis. In this work, we extend its results to kinematically represent the SRS skin (i.e., robot boundaries where we implement contact points). We parameterize the SRS skin at a radius, $r_{i}$ from the SRS neutral axis with an angular offset $ \sigma_{i}\in\left[0,2\pi\right] $ as shown in Fig.~\ref{fig:Fig3_SchematicDiagram}B. Considering any $i^{th}$ section ($i\in \{1,2,3\}$) of the SRS schematic shown in Fig.~\ref{fig:Fig3_SchematicDiagram}A,  
	the homogeneous transformation matrix (HTM) at any point on the skin of an SRS section, $\mathbf{T_{i}}\in \mathbb{SE}\left(3\right)$ can be derived as
	\begin{align}
		\mathbf{T}_{i}\left(\boldsymbol{q}_{i},\xi_{i}\right) & =\left[\begin{array}{cc}
			\mathbf{R}_{i}\left(\boldsymbol{q}_{i},\xi_{i}\right) & \mathbf{p}_{i}\left(\boldsymbol{q}_{i},\xi_{i}\right)\\
			\mathit{\boldsymbol{0}} & 1
		\end{array}\right]\cdots\nonumber \\
		& \qquad\left[\begin{array}{cc}
			\mathbf{R}_{z}\left(\sigma_{i}\right) & \mathcal{\mathit{\boldsymbol{0}}}\\
			\mathit{\boldsymbol{0}} & 1
		\end{array}\right]\left[\begin{array}{cc}
			\boldsymbol{1} & \mathbf{p}_{x}\left(r_{i}\right)\\
			\mathit{\boldsymbol{0}} & 1
		\end{array}\right]\label{eq:ith_kin}
	\end{align}
	where $\mathbf{R}_{i}\in\mathbb{SO} \left(3\right)$ and $\mathbf{p}_{i}\!\in\mathbb{R}^{3}$ denote the rotational matrix and the position vector, respectively. $\xi_{i}\in\left[0,1\right]$ is a scalar that defines points along a section, where the values 0 and 1 correspond to the origin and tip of a section, respectively. $\mathbf{R}_{z}\in\mathbb{SO}\left(3\right)$ is the rotation matrix about the $+Z$ and $\mathbf{p}_{x}\in\mathbb{R}^{3}$ -- translation matrix along the $+X$ of $\left\{ O_{i}^{\prime}\right\} $ is used to express modal kinematics on the robot skin. 
	
	By integrating a floating-base coordinate frame, $\mathbf{T}_{b}\in\mathbb{SE} \left(3\right)$ with~\eqref{eq:ith_kin}, the complete kinematic model of the 3-section SRS was derived as 
	\vspace{-1mm}
	\begin{align}
		\mathbf{T}\left(\boldsymbol{q}_{b},{\boldsymbol{q}_r},{\xi}\right) & =\mathbf{T}_{b}\left(\boldsymbol{q}_{b}\right)\prod_{i=1}^{3}\mathbf{T}_{i}\left({\boldsymbol{q}_{i}},\xi_{i}\right)\nonumber \\
		& =\left[\begin{array}{cc}
			\mathbf{R}\left(\boldsymbol{q}_{b},{\boldsymbol{q}_r},\xi\right) & \mathbf{p}\left(\boldsymbol{q}_{b},{\boldsymbol{q}_r},\xi\right)\\
			\mathit{\mathbf{0}} & 1
		\end{array}\right]\label{eq:complete_kin}
	\end{align}
	where $\boldsymbol{q}_{b}=\left[x_{b},y_{b},z_{b},\alpha,\beta,\gamma\right]\in\mathbb{R}^{6}$
	denotes the floating-base coordinate system parameters with $\left[x_{b},y_{b},z_{b}\right]$ and $\left[\alpha,\beta,\gamma\right]$ defining the linear and angular displacement of $\left\{ O_b\right\}$ relative to 
	$\left\{O\right\}$ (Fig.~\ref{fig:Fig3_SchematicDiagram}B). The vector ${\boldsymbol{q}_r}=[\boldsymbol{q}_{1},\boldsymbol{q}_{2},\boldsymbol{q}_{3}]\in\mathbb{R}^{9}$ defines the actuated jointspace of the SRS 
	with ${\xi}=\left[0,3\right]\in\mathbb{R}$ is a scalar that represents points along the SRS neutral axis. 
	By combining $\boldsymbol{q}_{b}$ and $\boldsymbol{q}_r$, we define the complete floating-base jointspace vector, $\boldsymbol{q}=\left[\boldsymbol{q}_{b},{\boldsymbol{q}_r}\right]\in\mathbb{R}^{15}$.

	\begin{figure}[tb] 
		\centering
		\includegraphics[width=1\linewidth]{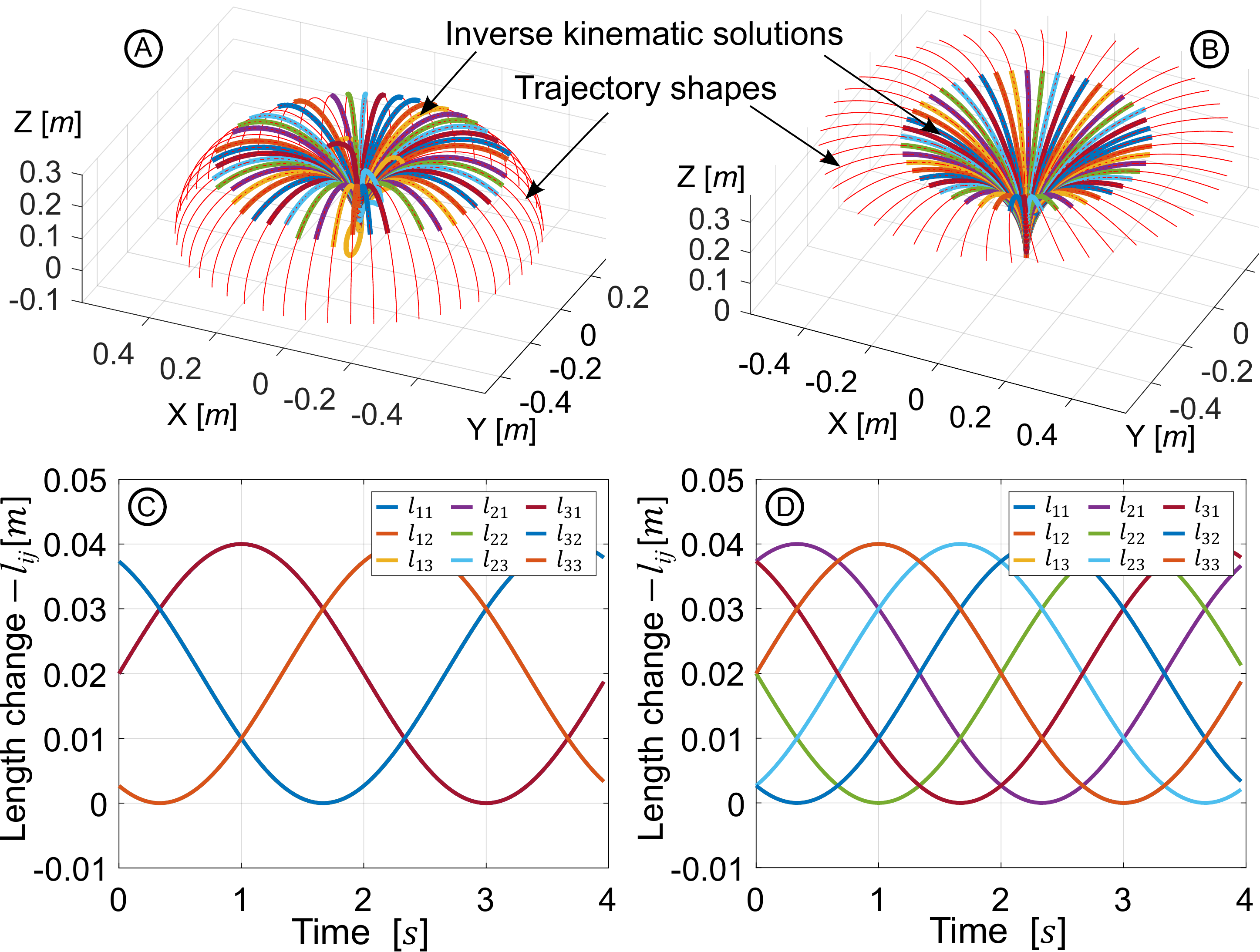}
		\caption{Trajectory curves of (A) planar \& (B) spatial rolling gaits relative to the robot origin. Jointspace trajectories of (C) planar \& (D) spatial rolling gaits during a gait cycle. In planar rolling, joint variables overlap each other.} 
		\label{fig:Fig5_MathematicalCurvesLengthVariable} 
	\end{figure}

	\subsection{Review of Trajectory Generation\label{subsec:Trajectory-Generation}}

	The SRS kinematic model is given in Sec.~\ref{subsec:SRS-Kinematic-Model}
	is used to derive the following locomotion trajectories. 
	We consider two trajectories.
	First is 
	planar rolling 
	where SRS sections share their bending on the same plane.
	The second is spatial rolling. In this work, we extend the planar rolling reported in~\cite{arachchige2021soft} to introduce this new locomotion gait where	
	each SRS section bends on separate bending planes (relative to $\{O_b\}$) creating a spatial bending pattern similar to a helix. 
	It is achieved by applying the jointspace trajectories computed for the planar rolling and actuating the adjacent SRS bending sections with an added constant angular phase shift, $\varphi = \frac{\pi}{3}$. Figs.~\ref{fig:Fig4_TrajectoryGeneration}A and~\ref{fig:Fig4_TrajectoryGeneration}B visualize these rolling patterns relative to SRS origin. 
	The trajectory generation procedure includes the following steps; i) identifying desired locomotion gait trajectory with respect to the global coordinate frame, ii) discretizing a gait trajectory cycle, iii) projecting the gait curve at discretized locations to the robot coordinate system, and finally, iv) employing an optimization-based inverse kinematic approach to obtain a joint space trajectory. 
	Readers are referred to our previous work~\cite{arachchige2021soft} for more details pertaining to these steps.

	Following the same steps, derived trajectory curves of planar and spatial rolling gaits relative to the robot coordinate frame are shown in Figs.~\ref{fig:Fig5_MathematicalCurvesLengthVariable}A and~\ref{fig:Fig5_MathematicalCurvesLengthVariable}B, respectively. Therein, thin red lines show projected trajectory curves onto the robot origin and thick multi-color lines show matched SRS shapes (Refer to Sec III in~\cite{arachchige2021soft} for more details).
	Correspondingly, obtained jointspace trajectories for planar and spatial rolling gaits for a period of $4~s$ are shown in Figs.~\ref{fig:Fig5_MathematicalCurvesLengthVariable}C and~\ref{fig:Fig5_MathematicalCurvesLengthVariable}D, respectively. They are applied to validate the dynamic model in Sec.~\ref{sec:Dynamic_Model_Validation}.
	Note that, in planar rolling, all sections operate without a phase shift ($\varphi=0$). Hence, joint variables in each section overlap each other as visualized in Fig.~\ref{fig:Fig5_MathematicalCurvesLengthVariable}C. On the other hand, joint variables in spatial rolling operate with a phase shift ($\varphi=\frac{\pi}{3}$) as visualized in Fig.~\ref{fig:Fig5_MathematicalCurvesLengthVariable}D.
	
	\begin{figure}[tb] 
		\centering
		\includegraphics[width=0.7\linewidth]{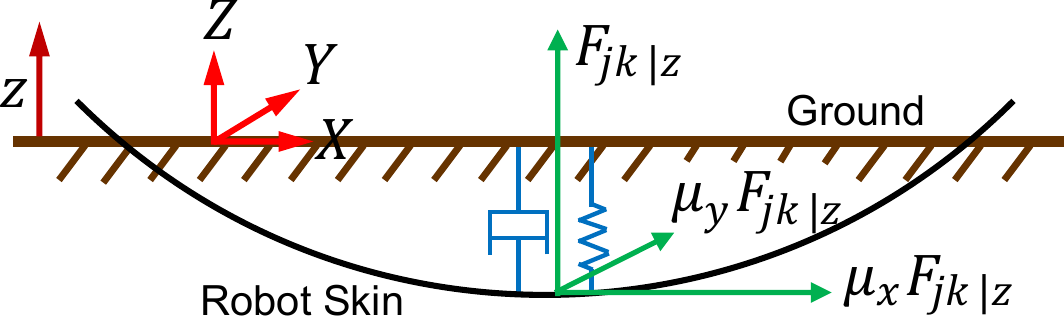}
		\caption{Contact dynamic model (Condition for the ground contact: $z<0$). The displacements are exaggerated for visualization.}
		\label{fig:Fig6_ContactDynamicModel} 
	\end{figure}
	
	\section{Dynamic Modeling \label{sec:Dynamic_Modeling}}
	
	The SRS dynamic modeling includes two components; robot-ground contact dynamics which includes the distributed contact dynamics along the robot snakeskin and complete SRS dynamics which presents the equations of motion (EoM) of the SRS. 
	
	\subsection{Robot-Ground Contact Dynamics\label{subsec:Contact-Dynamics}}
	
	We extend the dynamic model derived for variable--length multisection continuum robotic manipulators reported in~\cite{godage2016dynamics} to model the SRS considered here. However, the dynamic model cannot be directly utilized for modeling the SRS. Because, therein, the continuum manipulator has a fixed base whereas SRS achieves locomotion via different gaits. In addition, the model does not handle external forces. But, similar to snakes, the SRS achieves locomotion when its skin-like outer layer makes distributed contact with a surface and generates reaction forces via friction. 	
	
	Hence, we modify the model reported in~\cite{godage2016dynamics} to include a floating coordinate frame to support locomotion modeling and distributed contact dynamics and use it as a snake. The utility of moving coordinate systems on soft locomoting robots has been explored in~\cite{godage2012path,godage2012locomotion}. However, implementing continuous contact dynamics is computationally inefficient. We use a discrete model with an array of finely distributed contact points defined along the periphery of the SRS by introducing two parameters, $\mathbf{\xi} \in \left[0,3\right]$ and $\sigma_{i} \in \left[0,2\pi\right]$ which discretize the SRS surface into 31 points axially and 10 points radially as shown in~\eqref{eq:ith_kin} and~\eqref{eq:complete_kin}. This results in 310 contact points on the outer layer of the SRS. We compute the reaction forces of those points when contact conditions are met using a spring-damper ground model (Fig.~\ref{fig:Fig6_ContactDynamicModel}). We define a ground contact condition as when the z coordinate of a contact point with respect to $\{O\}$ is negative, i.e., $z<0$,~\cite{marhefka1996simulation}. As long as this condition is met, the reaction forces are continuously computed and added to the SRS dynamic model as follows. Without losing generality, let the $z$ component of the ground reaction force at any contact point (defined by $\xi_{j}$ and $\sigma_{k}$), be $F_{jk | z}$ and given by
	\vspace{-1mm}
	\begin{align}
		F_{jk | z} & = -\frac{1}{2}\left(1-\text{sign}\left(z\right)\right)\left(K_{g}z+B_{g}\dot{z}\right) \label{eq:z_reaction_force}
	\end{align}
	where $K_{g}$ and $B_{g}$ are the ground stiffness and damping coefficients, respectively. 
	
	Here, we assume that the ground stiffness is sufficiently large (i.e., high $K_{g}$) and thus $z$ is negligible such that our point-contact model is valid. 
	To achieve locomotion, there must be net reaction forces on the $X-Y$ plane (Fig.~\ref{fig:Fig6_ContactDynamicModel}). 
	From standard ground friction models, we can compute the complete reaction force $F_{jk}$ as
	\vspace{-1mm}
	\begin{align}
		\mathbf{F}_{jk} & =F_{jk|z}\left[
		\begin{array}{ccc}
			\mu_{x}\text{sign}\left( \dot{x} \right)  & \mu_{y}\text{sign}\left( \dot{y} \right) & 1
		\end{array}
		\right]^{T}
		\label{eq:total_reaction_force}
	\end{align}
	where $\mu_x$ and $\mu_y$ are the static reaction coefficients in the $X$ and $Y$ directions
	respectively.
	
	\begin{figure}[tb] 
		\centering
		\includegraphics[width=0.85\linewidth]{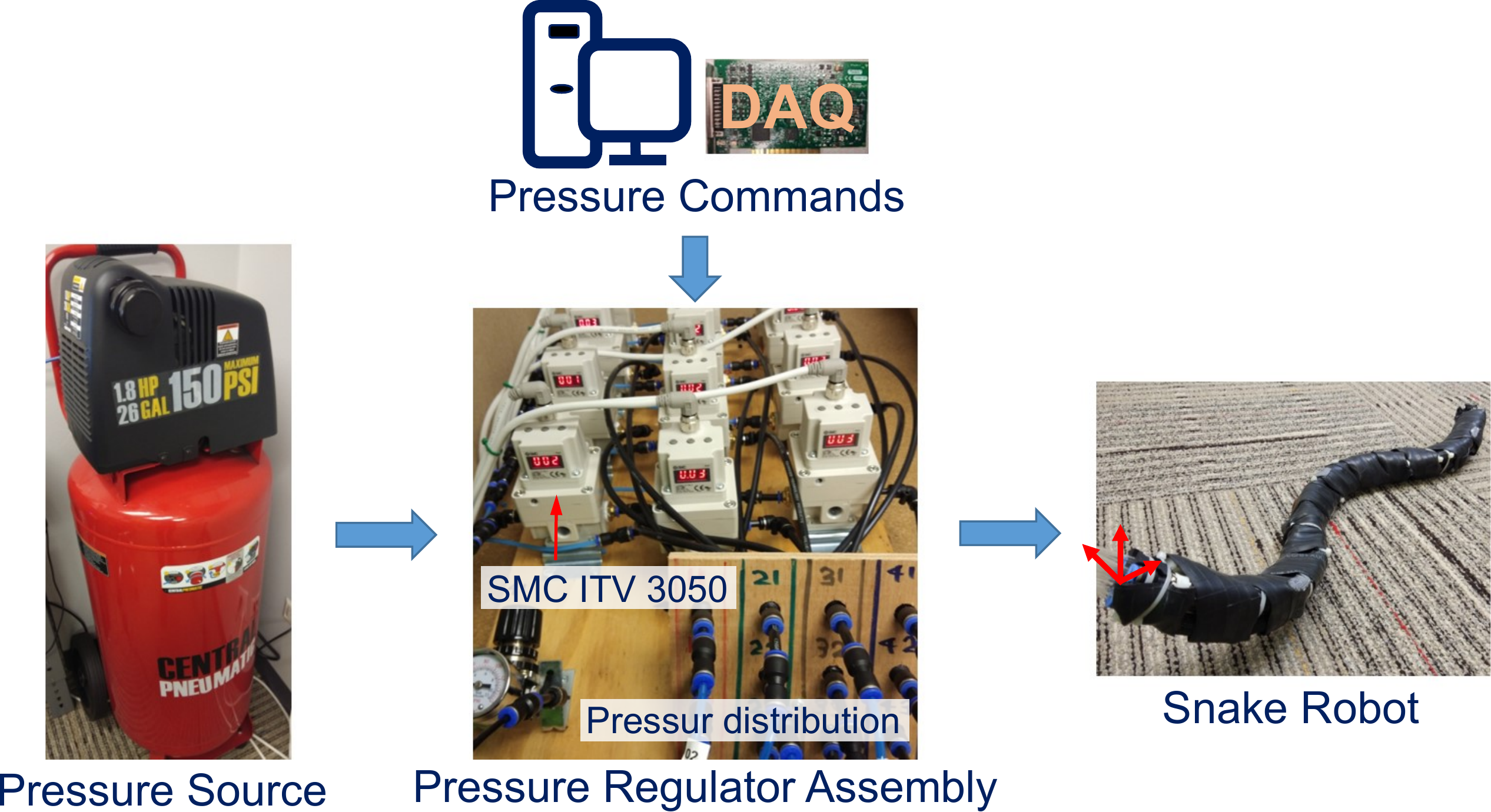}
		\caption{ Robot experimental setup.}
		\label{fig:Fig7_ExperimentalSetup} 
	\end{figure}

	\subsection{Complete Soft Robotic Snake Dynamics\label{subsec:Robot_Dynamics}}
	
	We assume that an SRS section is made of a set of infinitesimally thin slices with constant mass and uniform linear density. Using the floating base kinematics derived in~\eqref{eq:complete_kin}, we calculate the kinetic and potential energies of a thin slice and integrate them to find the total energies of bending sections. We then apply the Lagrangian mechanics-based recursive computation scheme proposed in~\cite{godage2016dynamics} to derive the EoM of the SRS as 
	\vspace{-1mm}
	\begin{align}
		\mathbf{M}\ddot{q}+\left( \mathbf{C+D}\right) \dot{q}
		+\mathbf{G}& =\left[\begin{array}{c}
			0\\
			\boldsymbol{\tau}_{e}
		\end{array}\right]+\sum_{j\in\xi,k\in\sigma}\mathbf{J}_{jk}^{T}\mathbf{F}_{jk}\label{eq:simulatino_EoM}
	\end{align}
	where $\mathbf{M}\in\mathbb{R}^{15\times15}$ is the generalized inertia matrix, $\mathbf{C}\in\mathbb{R}^{15\times15}$ is the centrifugal and Coriolis force matrix, $\mathbf{D} \in\mathbb{R}^{15\times15}$ is the damping force matrix, and $\mathbf{G}\in\mathbb{R}^{15}$ is the conservative force vector. Here, $\boldsymbol{\tau}_{e} \in \mathbb{R}^{9}$
 defines the pressure force vector and $\mathbf{J}_{jk}$ defines the Jacobian, which maps $\mathbf{F}_{jk}$ into jointspace $\boldsymbol{q}$. Note that, this dynamic model does not consider the hysteretic effects as it is negligible compared to the damping effect~\cite{godage2016dynamics}.
	
	\begin{figure}[tb] 
		\centering
		\includegraphics[width=1\linewidth]{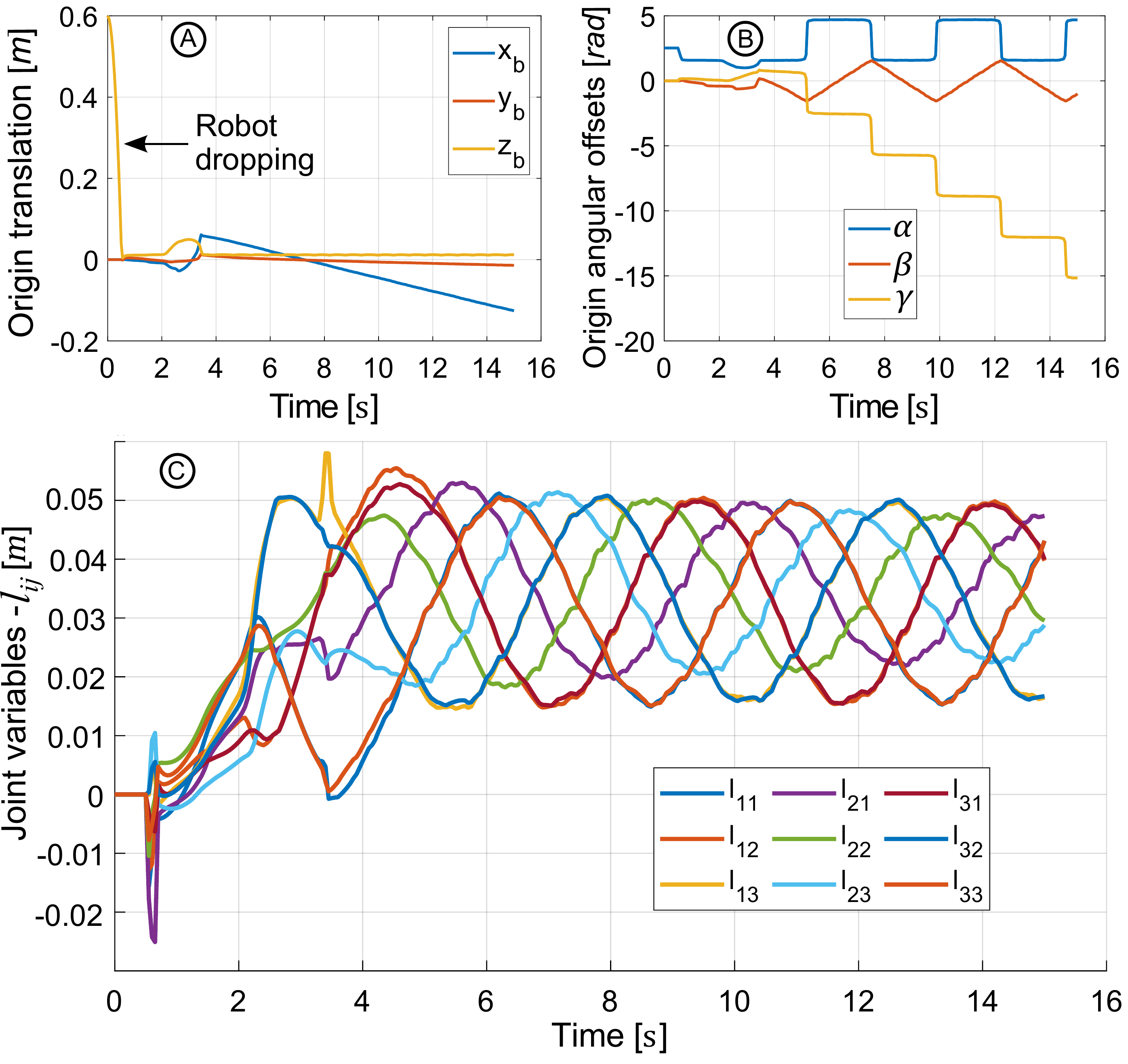}
		\caption{Dynamic model simulation outputs for spatial rolling -- (A) Position \& (B) Orientation changes of SRS origin. C) Joint variables of complete SRS.} 
		\label{fig:Fig8_DynamicModelSimulationOutput} 
	\end{figure}
	
	\begin{figure*}[tb] 
		\centering
		\includegraphics[width=1\linewidth]{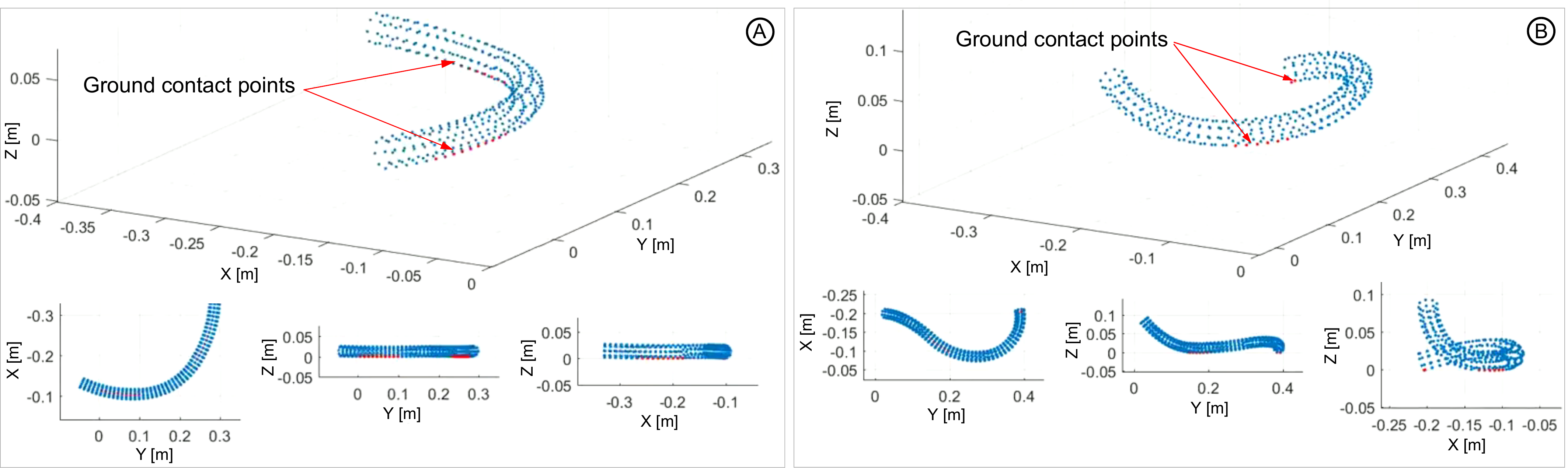}
		\caption{Dynamic model implementation in an SRS numerical model and simulated locomotion gaits -- (A) Planar rolling, (B) Spatial rolling.} 
		\label{fig:Fig9_DynamicModelSimulationVisualization}
	\end{figure*}
	
	In the recursive formulation employed here, the terms in~\eqref{eq:simulatino_EoM} can be separated into contributions from each $i^{th}$ SRS bending section. For instance, 
	the generalized inertia matrix
	can be written as $\mathbf{M}=\sum_{i=1}^{3}\mathbf{M}_{i}$ where $\mathbf{M}_{i}$ is the inertia matrix of the $i^{th}$ SRS bending section given by
	\vspace{-1mm}
	\begin{align}
		\left [{\mathbf{M}_{i}} \right ]_{uv} & =m_i \int_{\xi_i} \frac{\partial \boldsymbol{p}_i^T}{\partial \boldsymbol{q}_u}  \frac{\partial \boldsymbol{p}_i}{\partial \boldsymbol{q}_v} 
		\label{eq:M_Vw}		 
	\end{align}
	where $m_i$ is the mass of the $i^{th}$ section, 
	and $\{u,v\} \in \mathbb{Z}^{+} \land [1,\cdots,15]$ denotes the matrix element index. 
	
	For any $i^{th}$ SRS section, the elements of the centrifugal and Coriolis force matrix,  $\mathbf{C}_{i} \in \mathbb{R}^{15 \times 15}$ can be derived from the partial derivatives of $\mathbf{M}_i$ as reported in~\cite{godage2016dynamics},
	\vspace{-1mm}
	\begin{align}
		[\mathbf{C}_{i}]_{vu} & =\sum_{h=1}^{15}\mathbf\Gamma_{vuh}\left ( \mathbf{M}_{i} \right )\dot{q}_{h}, \mbox{and} 
		\label{eq:C_i}
		\\
		\mathbf\Gamma_{vuh}(\mathbf{M}_i) & =  \frac{1}{2}\left ( \frac{\partial [\mathbf{M}_i]_{vu}}{\partial q_h}+ \frac{\partial [\mathbf{M}_i]_{vh}}{\partial q_u}-\frac{\partial [\mathbf{M}_i]_{hu}}{\partial q_v} \right)
	\end{align}
	
	Recursively, $\mathbf{C}=\sum\nolimits_{i=1}^{3}\mathbf{C}_i$ gives the complete centrifugal and Coriolis force matrix.
	
	Damping force matrix, $\mathbf{D}_i$ 
	for any $i^{th}$ SRS section can be written as a diagonal matrix such that, $\mathbf{D}_i= diag([D_{i1}, D_{i2}, D_{i3}]) \in \mathbb{R}^{3 \times 3}$. Then, $\mathbf{D}=\sum\nolimits_{i=1}^{3}\mathbf{D}_i$ gives the complete damping force matrix.
	
	The conservative force vector for any $i^{th}$ SRS section, $\mathbf{G}_i$ can be written as
	\begin{align}
		\mathbf{G}_{i}= \mathbf{K}_i q_i+m_i\int_{\xi_i}  \frac{\partial \boldsymbol{p}_i^T}{\partial q_i}  \boldsymbol{g}
		\label{eq:G_i}
	\end{align}
	where $\mathbf{K}_i$ is the elastic stiffness coefficient matrix of any $i^{th}$ SRS section. It can be written as a diagonal matrix such that, $\mathbf{K}_i= diag([K_{i1}, K_{i2}, K_{i3}]) \in \mathbb{R}^{3 \times 3}$.
	$\boldsymbol{g}=[0,0,g]^T$ is the gravitational acceleration vector. 
	
	Employing the recursive approach, the complete conservative force vector can be written as $\mathbf{G}=\sum\nolimits_{i=1}^{3}\mathbf{G}_i$.
	The readers are referred to~\cite{godage2016dynamics} for a detailed derivation of the EoM.

	\section{Dynamic Model Validation\label{sec:Dynamic_Model_Validation}}
	
	We carry out the dynamic model validation in three steps. In the first step, we 
	implement the dynamic model as a numerical model and apply locomotion trajectories derived in Sec.~\ref{subsec:Trajectory-Generation}
    and 
	simulate them in a contact-enabled simulation environment. The second step
	involes the application of the same locomotion trajectories tested in the first step to the actual SRS hardware and experimentally 
	evaluated for gait replications. In the last step, we
	qualitatively and quantitatively compare the numerical model outputs with experimental results and validate the dynamic model.
	
	\subsection{Dynamic Model Simulation\label{subsec:Dynamic-Model-Simulation}}
	
	We implemented the SRS dynamic model derived in Sec.~\ref{subsec:Robot_Dynamics} as a numerical model
 and provided jointspace trajectories (Figs.~\ref{fig:Fig5_MathematicalCurvesLengthVariable}C and~\ref{fig:Fig5_MathematicalCurvesLengthVariable}D) as inputs to test for gait simulations. 
The numerical model was implemented in the MATLAB 2021a programming environment, and simulations were recorded. MATLAB's ODE15 solver is selected for solving the~\eqref{eq:simulatino_EoM} due to the stiff nature of the complex, high-DoF dynamic systems such as the one presented here. 
 Herein, jointspace trajectories (i.e., length changes) are converted into pressure trajectories and then applied as force inputs ($= pressure \times sectional\ area\ of\ PMAs$)
 (see Sec.~\ref{subsec:testing_rolling_gaits}). We approximated PMA elastic stiffness limiting values as $K_{i1}=1900~Nm^{-1}\, \forall i \in \{1,2,3\}$ (rounded to the nearest 100) and the damping coefficients as $D_{i1}=90~Nm^{-1}s\, \forall i \in \{1,2,3\}$ (rounded to the nearest 10) following an experimental procedure similar to the one proposed in~\cite{godage2016dynamics}. We assumed that the robot is actuating on a carpeted floor (Fig.~\ref{fig:Fig1_IntroductionImage}) that has uniform friction. We experimentally approximated the ground stiffness as $K_g=1000~Nm^{-1}$, damping as $B_g=130~Nm^{-1}s$ and, static frictional coefficients as ${\mu}_x=0.6,\ {\mu}_y=0.2$.
 The gravitational acceleration was set as $9.81~ms^{-2}$. 
	
	\begin{figure*}[tb] 
		\centering
		\includegraphics[width=1\textwidth, height=0.6\columnwidth]{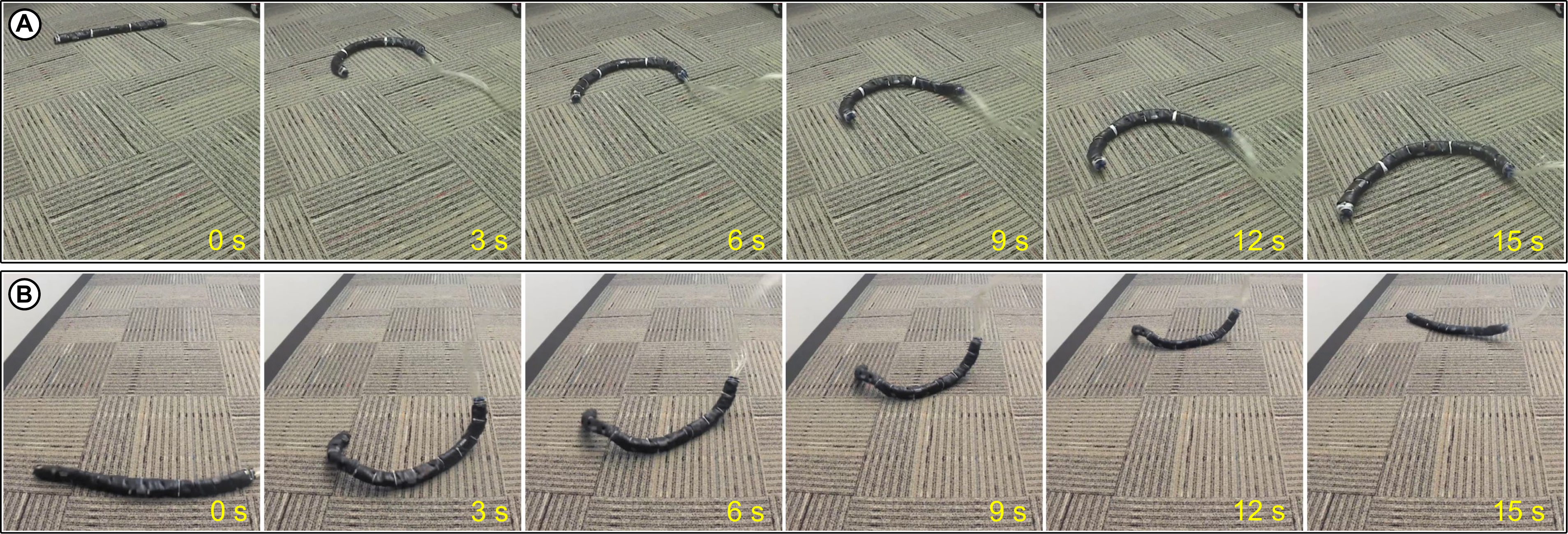}
		\caption{The SRS pose progression for (A) Planar and (B) Spatial rolling gaits at $3~bar-0.50~Hz$.}
		\label{fig:Fig10_RollingSpirallingResults} 
	\end{figure*}
	
	\begin{figure}[tb] 
		\centering
		\includegraphics[width=1\linewidth]{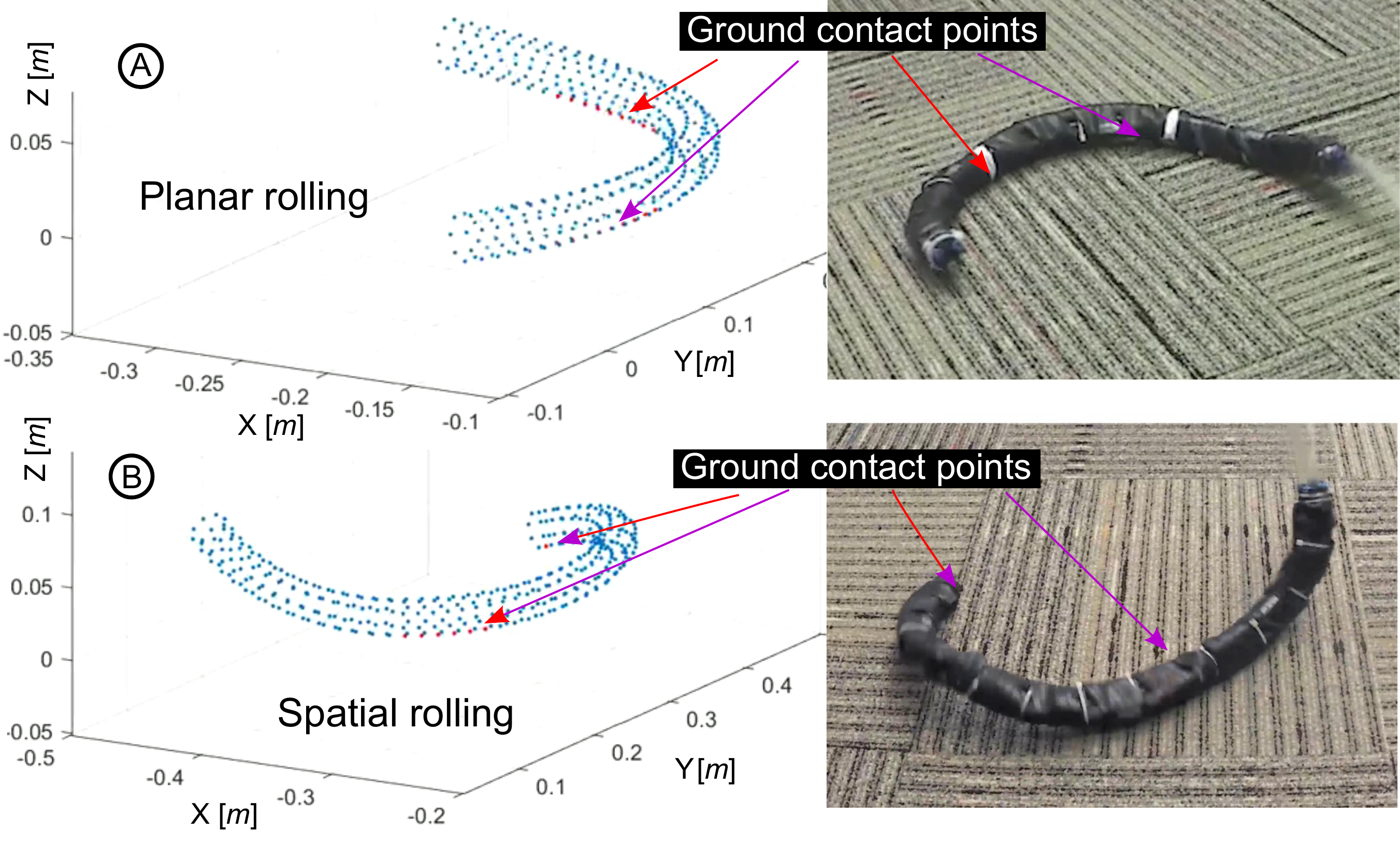}
		\caption{Ground contacts in dynamic model and SRS prototype. Contact points in the numerical model are shown by red color dots.}
		\label{fig:Fig11_GroundContactComparison} 
	\end{figure}
	
	\begin{figure*}[tb] 
		\centering
		\includegraphics[width=0.95\linewidth]{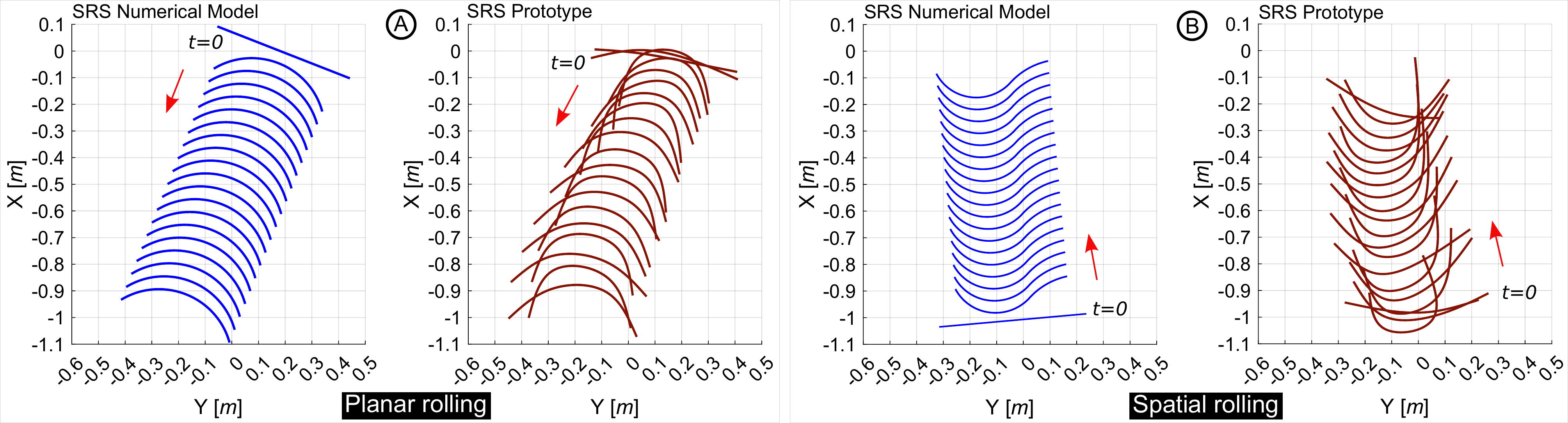}
		\caption{Locomotion tracking of numerical model vs SRS prototype. Herein, the SRS neutral axis has been projected to $X-Y$ plane of $O$.}
		\label{fig:Fig12_LocomotionTracking} 
	\end{figure*}
	
	\subsection{Experimental Setup\label{subsec:Experimental-Setup}}
	
	The experimental setup prepared for the SRS testing is shown in Fig.~\ref{fig:Fig7_ExperimentalSetup}. We use a constant pressure source (a compressor with $8~bar$) to supply air pressure to digital proportional pressure regulators (SMC ITV3050) and then regulate pressure to individual PMAs of the SRS -- 
	nine regulators are used for nine PMAs in three robot bending sections. Each regulator is independently controlled by a $0-10~V$ analog input voltage signal provided through a data acquisition (DAQ) card (National Instruments PCI-6221). The DAQ card is installed in a host computer, and control signals are generated using a MATLAB Simulink Desktop Real-Time model.

	\subsection{Numerical Testing}\label{subsub:numerical_testing}
	In the simulation, first, we dropped the robot from a known height ($0.6~m$) and then engaged the SRS in rolling trajectories. The dropping test was carried out to evaluate the contact dynamic model as explained in the subsequent text.
 The displacement of the origin of $\{O_b\}$ is depicted in Figs.~\ref{fig:Fig8_DynamicModelSimulationOutput}A and~\ref{fig:Fig8_DynamicModelSimulationOutput}B. The figures visualize position ($x,y,z$) and orientation ($\alpha,\beta,\gamma$) changes of the robot coordinate system origin (i.e., floating-base parameter changes) during a simulation period of $15~s$ with respect to spatial rolling. The joint variable output ($l_{ij}$) is presented in Fig.~\ref{fig:Fig8_DynamicModelSimulationOutput}C. It closely resembles the trajectory input shown in Fig.~\ref{fig:Fig5_MathematicalCurvesLengthVariable}D. Initial drop
	($0-2~s$ in Fig.~\ref{fig:Fig8_DynamicModelSimulationOutput}A) helps us examine the validity of the contact dynamic model stated in Sec.~\ref{subsec:Contact-Dynamics} as follows. After the drop, it is clear that $z$ stops at $z=0$, assuring ground contact conditions. Additionally, Fig.~\ref{fig:Fig8_DynamicModelSimulationOutput}A proves that, throughout the simulation, the robot stays
	above the ground. We recorded simulation data at 
	$30~Hz$ sampling rate to ensure a smooth approximation of jointspace variable changes
	in simulation videos.
	Two separate movie frames of planar and spatial rolling simulations are presented in Fig.~\ref{fig:Fig9_DynamicModelSimulationVisualization}. Readers are referred to the accompanying multimedia file (\url {https://youtu.be/V_RNFEiVXlw}) to see the complete simulations. The simulation outputs show that the dynamic model replicates the desired gaits well, demonstrating the intended operation of ground contact dynamics. 
	
	\subsection{Prototype Testing \label{subsec:testing_rolling_gaits}}
    The locomotion gaits are tested on the SRS prototype to compare and validate the results obtained in numerical testing. 
	We tested the SRS prototype for planar and spatial rolling gaits on a carpeted floor with uniform friction. The jointspace trajectories are length changes of PMAs, and they must be actuated in order to obtain locomotion from the SRS prototype. The length changes of PMAs are a function of input pressures. Therefore, we adopted the approach used in~\cite{arachchige2021soft} to establish the length-to-pressure mapping and supply pressure inputs accordingly. We applied the same jointspace trajectories used in dynamic model simulations (i.e., length parameters in Figs.~\ref{fig:Fig5_MathematicalCurvesLengthVariable}C and~\ref{fig:Fig5_MathematicalCurvesLengthVariable}D) to obtain pressure trajectories and actuate the SRS prototype. The locomotion trajectories were tested for $15~s$ at a maximum supply pressure of $3~bar$ and frequency $0.50~Hz$, which are consistent with the simulation inputs. A $3~bar$ pressure ceiling was used based on PMAs' ability to achieve the required SRS deformation. The frequency range was chosen based on the operational bandwidth of PMAs to obtain meaningful locomotion. The SRS testing was video captured using a fixed camera station. The locomotion progression of the SRS during planar and spatial rolling gaits is shown in Figs.~\ref{fig:Fig10_RollingSpirallingResults}A and~\ref{fig:Fig10_RollingSpirallingResults}B respectively. Our multimedia file (\url {https://youtu.be/V_RNFEiVXlw}) shows the complete results of these experiments. The results show that similar to the numerical testing, the SRS prototype replicated the desired locomotion trajectories well on the carpeted floor. 
	
	\subsection{Discussion\label{subsec:ModelAssessments}}

 \begin{table}[b]
		\setlength{\tabcolsep}{5pt} %
		\centering
		\caption{numerical and experimental model outputs.}
		\label{Table:numericalandexperimental}
		\begin{tabular}{|c|c|c|c|c|} 
			\hline
			\multirow{3}{*}{\begin{tabular}[c]{@{}c@{}}\\Model\end{tabular}} & \multicolumn{4}{c|}{\begin{tabular}[c]{@{}c@{}}Travelling velocity \\ ($cms^{-1}$)\end{tabular}} \\ 
			\cline{2-5}
			& \multicolumn{2}{l|}{Planar rolling} & \multicolumn{2}{l|}{Spatial rolling} \\ 
			\cline{2-5}
			& $V_x$ & $V_y$ & $V_x$ & $V_y$ \\ 
			\hline
			\multicolumn{1}{|l|}{SRS Numerical Model -- $V_N$} & 3.51 & 9.39 & 0.67 & 7.77 \\ 
			\hline
			\multicolumn{1}{|l|}{SRS Prototype -- $V_P$} & 3.31 & 9.01 & 0.61 & 7.12  \\
			\hline
			\multicolumn{1}{|l|}{Error [$\%$] $=\frac{V_N-V_P}{V_N}\times100\ \%$} & 5.70 & 4.05 & 8.96 & 8.37 \\
			\hline
		\end{tabular}
	\end{table}
	
	The Figs.~\ref{fig:Fig11_GroundContactComparison}A and~\ref{fig:Fig11_GroundContactComparison}B show respective contact point mapping between dynamic model simulations and SRS prototype testing. Similar to the dynamic model simulations, we observed that the SRS prototype could successfully replicate two rolling gaits. Note that, here we applied the same jointspace trajectories (in Figs.~\ref{fig:Fig5_MathematicalCurvesLengthVariable}C and~\ref{fig:Fig5_MathematicalCurvesLengthVariable}D) to the numerical model and the SRS prototype. Hence, the replication of closely resembled locomotion patterns with contact points in both models qualitatively confirms the validity of the proposed dynamic model. 
	
	We tracked numerical and experimental model outputs to quantify and compare the dynamic model performance. Refer to Figs.~\ref{fig:Fig12_LocomotionTracking}A and~\ref{fig:Fig12_LocomotionTracking}B show the captured $X-Y$ displacement of the SRS during planar and spatial rolling gait replications. The left in each figure shows the $X-Y$ displacement of the numerical model. The experimental displacement data shown on the right were captured using the image perspective projection method reported in~\cite{arachchige2021soft}. Based on displacement data, the calculated linear velocity components are presented in Table~\ref{Table:numericalandexperimental}. Results in Table~\ref{Table:numericalandexperimental} show that the velocity components of the SRS numerical model closely follow (Error,
 < 09~\%) its counterpart, i.e., the SRS prototype, thereby quantitatively verifying the proposed dynamic model. 
	
	\section{Conclusions}
	The SRSs lack spatial dynamic models that support distributed contact dynamics. We proposed a dynamic model with simplified point-based contact dynamics for a 3-section SRS in this work. The proposed dynamic model is capable of replicating spatial locomotion gaits. We presented a kinematic model of the SRS and used it to obtain jointspace trajectories for two types of locomotion gaits known as planar and spatial rolling. First, we implemented the SRS dynamic model as a numerical model and applied jointspace trajectories in a simulation environment. Next, we applied the same jointspace trajectories to the SRS prototype and experimentally tested the SRS's ability to replicate the intended two gaits. The SRS dynamic model and the prototype replicated gaits well. Both tests gave qualitatively and quantitatively consistent results, thus validating the proposed dynamic model. We intend to extend the dynamic model validation into other snake locomotion gaits in our future work. 
	
	\bibliographystyle{IEEEtran}
	\bibliography{refs}
	
\end{document}